\documentclass[runningheads]{llncs}
\usepackage{amsmath,amssymb}
\usepackage[T1]{fontenc}
\usepackage{graphicx,verbatim}
\begin{document}
\title{BrainSCL: \underline{S}ubtype-Guided \underline{C}ontrastive \underline{L}earning for Brain Disorder Diagnosis}
\titlerunning{BrainSCL for Brain Disorder Diagnosis}
%

\author{
Xiaolong Li\inst{1} \and
Guiliang Guo\inst{1} \and
Guangqi Wen\inst{2} \and
Peng Cao\inst{1} \and
Jinzhu Yang\inst{1} \and
Honglin Wu\inst{1} \and
Xiaoli Liu\inst{3} \and
Fei Wang\inst{4} \and
Osmar R. Zaiane\inst{5}
}

\authorrunning{X. Li et al.}

\institute{
Computer Science and Engineering, Northeastern University, Shenyang, China \\
\email{caopeng@cse.neu.edu.cn} \and
School of Computer Science and Artificial Intelligence, Shandong Normal University, Jinan, China \and
AiShiWeiLai AI Research, Beijing, China \and
Early Intervention Unit, Department of Psychiatry, Affiliated Nanjing Brain Hospital, Nanjing Medical University, Nanjing, China \\
\email{fei.wang@yale.edu} \and
Amii, University of Alberta, Edmonton, AB, Canada
}

\maketitle              
\begin{abstract}
Mental disorder populations exhibit pronounced heterogeneity — that is, the significant differences between samples — poses a significant challenge to the definition of positive pairs in contrastive learning.
To address this, we propose a subtype-guided contrastive learning framework that models patient heterogeneity as latent subtypes and incorporates them as structural priors to guide discriminative representation learning. 
Specifically, we construct multi-view representations by combining patients' clinical text with graph structure adaptively learned from BOLD signals, to uncover latent subtypes via unsupervised spectral clustering. 
A dual-level attention mechanism is proposed to construct prototypes for capturing stable subtype-specific connectivity patterns. 
We further propose a subtype-guided contrastive learning strategy that pulls samples toward their subtype prototype graph, reinforcing intra-subtype consistency for providing effective supervisory signals to improve model performance.
We evaluate our method on Major Depressive Disorder (MDD), Bipolar Disorder (BD), and Autism Spectrum Disorders (ASD). Experimental results confirm the effectiveness of subtype prototype graphs in guiding contrastive learning and demonstrate that the proposed approach outperforms state-of-the-art approaches. Our code is available at \url{https://anonymous.4open.science/r/BrainSCL-06D7}.

\keywords{Patient subtype \and Multi-view fusion \and Subtype-guided contrastive learning.}

\end{abstract}
\section{Introduction}
\begin{figure}[htbp]
\centering
\begin{tabular}{cc}
\begin{minipage}[t]{0.33\textwidth}
    \centering
    \includegraphics[width=\linewidth]{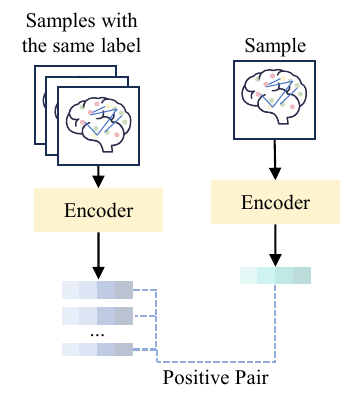}\\
    {\scriptsize (a) Conventional supervised CL}
\end{minipage}
&
\begin{minipage}[t]{0.57\textwidth}
    \centering
    \includegraphics[width=\linewidth]{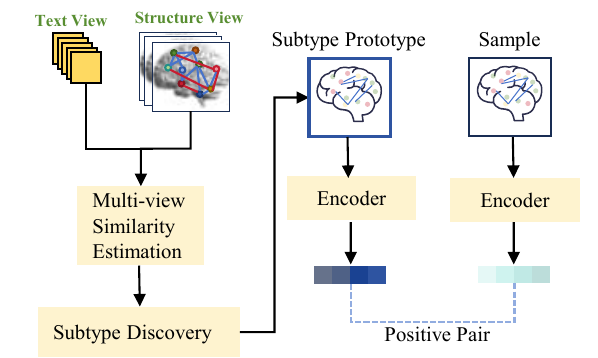}\\
    {\scriptsize (b) Subtype-guided CL}
\end{minipage}
\\
\begin{minipage}[t]{0.45\textwidth}
    \centering
    \includegraphics[width=\linewidth]{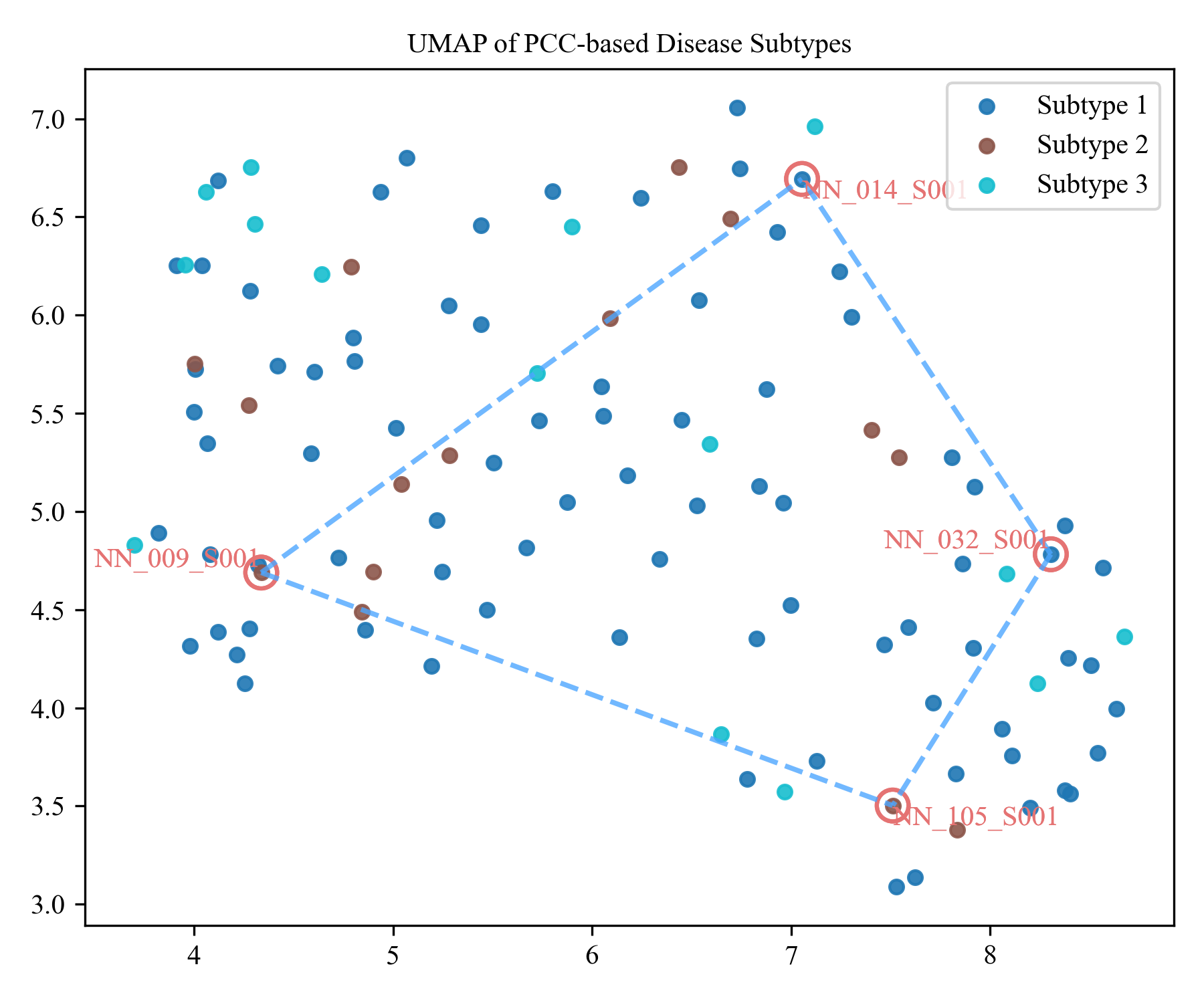}\\
    {\scriptsize (c) The positive pairs in space described by the original graph structure}
\end{minipage}
&
\begin{minipage}[t]{0.45\textwidth}
    \centering
    \includegraphics[width=\linewidth]{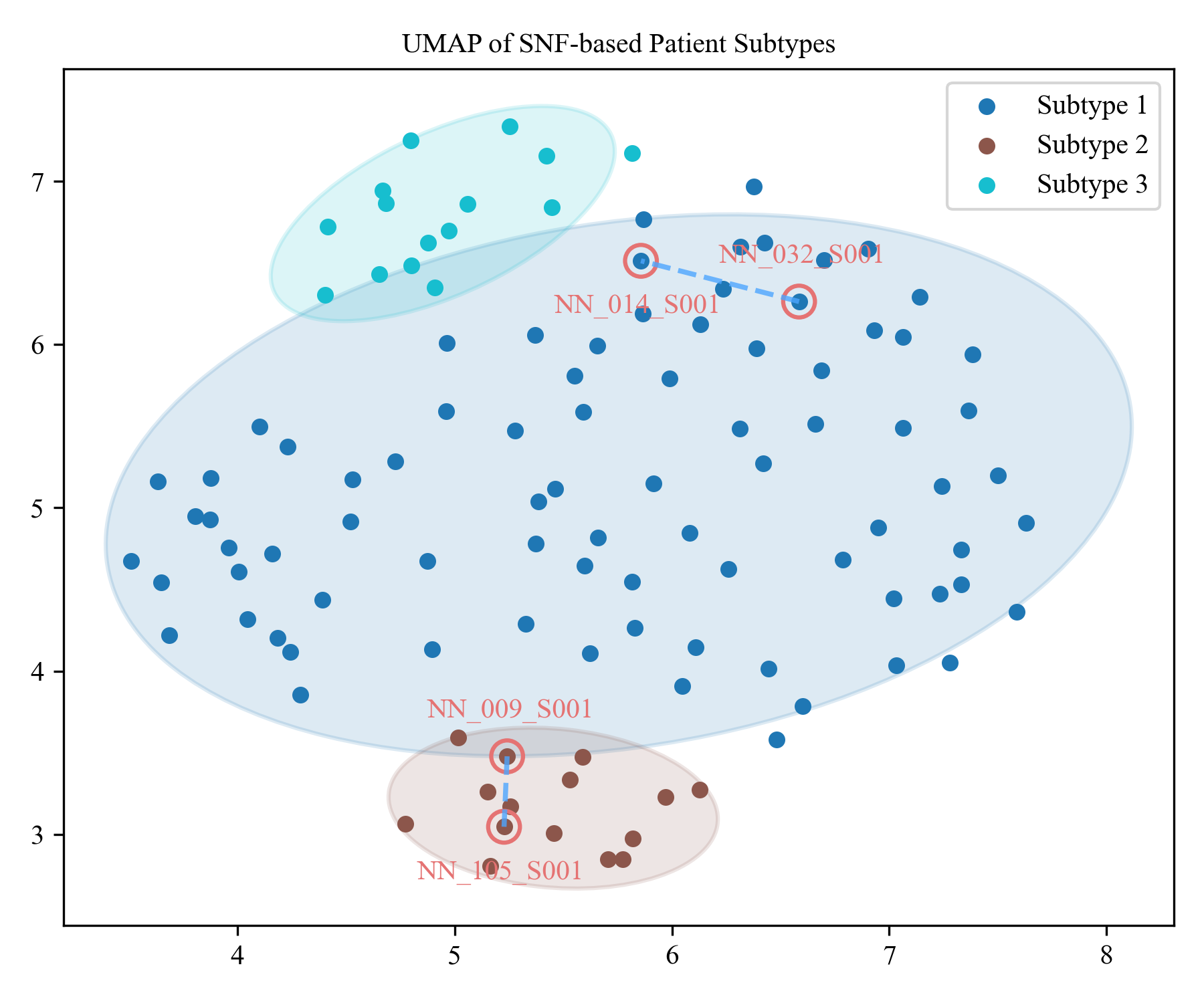}\\
    {\scriptsize (d) The subtype guided positive pairs in  space described by multi-view features}
\end{minipage}
\end{tabular}

\caption{Comparison of conventional supervised contrastive learning (CL) and our subtype-guided contrastive learning.}
\label{fig1}
\end{figure}

Objective diagnosis of psychiatric disorders has long been a central focus in neuroimaging research, and functional magnetic resonance imaging (fMRI)-based models have advanced automated diagnosis for disorders such as Major Depressive Disorder, Bipolar Disorder, and Autism Spectrum Disorders~\cite{Ahmed2025,Takahara2025}. 
In the context of limited disorder data, supervised contrastive learning (Fig.~\ref{fig1}(a)) provides effective supervisory signals to improve model performance~\cite{Li2024}. And the effectiveness of contrastive learning critically depends on the semantic reliability of positive pair construction~\cite{Adjeisah2024,Wang2022}. However,
extensive neuroimaging studies show that patients with the same clinical diagnosis can exhibit substantial connectivity variability in the brain network, reflecting the intrinsic heterogeneity of psychiatric disorders~\cite{Hu2024,Lawn2024,Secara2024}. This heterogeneity is further illustrated in Fig.~\ref{fig1}(c), which shows the 
highly variable connectivity patterns for patients.
The significant inter-sample variability within the same class violates the underlying assumption that positive pairs should be naturally similar, making the definition of valid positive pairs and the subsequent feature alignment particularly difficult. It causes the model to learn spurious correlations, ultimately hindering its ability to learn robust and generalizable representations.

Therefore, addressing the challenge of heterogeneity—specifically, rethinking what constitutes a meaningful positive pair—is crucial for effectively adapting contrastive learning to the medical domain.
To address this, we propose a subtype-guided contrastive learning framework (BrainSCL) in Fig.~\ref{fig1}(b) that constructs semantically meaningful positive supervision while explicitly accounting for latent inter-patient heterogeneity. BrainSCL identifies latent patient subtypes from complementary multi-views and summarizes each subtype with a prototype capturing connectivity patterns shared across its samples. These subtype-level prototypes serve as stable, biologically grounded guidance for defining positive pairs (Fig.~\ref{fig1}(d)).
Extensive experiments on multiple psychiatric disorder diagnosis tasks demonstrate that BrainSCL improves diagnostic performance, validating the effectiveness of incorporating patient subtypes into contrastive representation learning. The main contributions are summarized as follows:
\begin{enumerate}
    \item We design a multi-view patient subtype discovery strategy that integrates functional brain network structures with clinical text information, and summarizes each subtype with a representative prototype via a dual-level attention mechanism.
    \item We propose a subtype-guided contrastive learning strategy that provides principled, semantically grounded positive supervision while explicitly accounting for inter-patient heterogeneity in psychiatric disorder diagnosis.  
    \item We conduct systematic experiments on multiple psychiatric disorder diagnosis tasks, showing that our method outperforms state-of-the-art approaches and exhibits strong generalization. 
\end{enumerate}




\begin{figure}[htbp]
\centering
\includegraphics[width=0.9\textwidth]{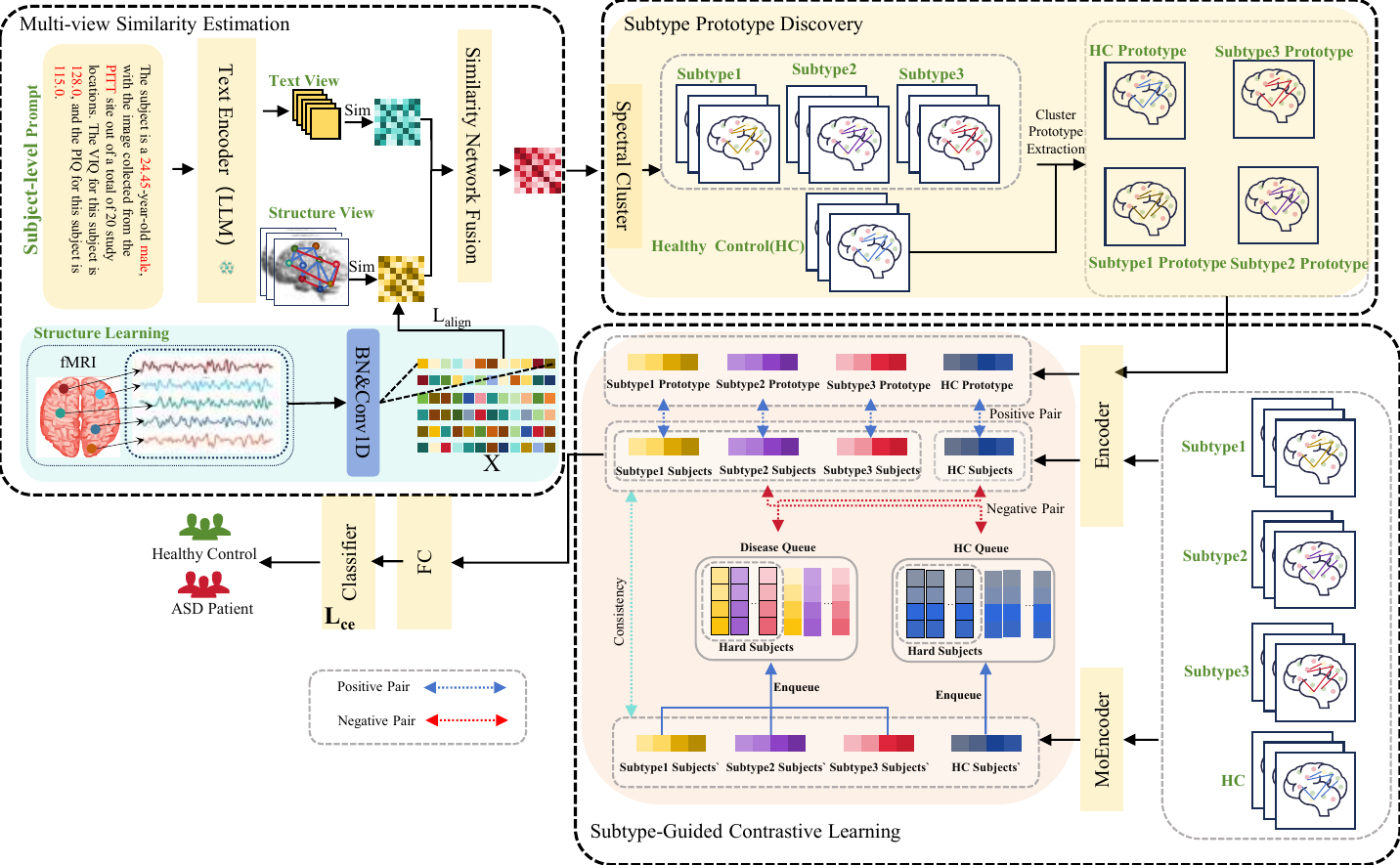}
\caption{Schematic of the proposed BrainSCL framework, which consists of three main modules: multi-view (text and graph structure) similarity estimation for generating a fused multi-view similarity matrix; 
subtype discovery for identifying latent subtypes and constructing subtype prototypes; 
subtype-guided contrastive learning for reinforcing intra-subtype consistency and enhancing discrimination between patients and healthy controls.}\label{fig3}
\end{figure}
\section{Method}
\subsection{Multi-view Similarity Estimation} To characterize latent patient subtypes, we construct a multi-view representation for each patient.
Directly computing functional connectivity via PCC easily introduces noise; therefore, we employ a trainable graph structure generator for constructing a structure view. For each ROI $r$, the BOLD time series ${S}_r \in \mathbb{R}^{T}$ is encoded via a convolutional encoder:  ${h}_r = f_{\theta}({S}_r)$, $r = 1, \ldots, M$, where $f_{\theta}$ denotes stacked 1D convolutional layers with batch normalization. 
Pairwise cosine similarities between node embeddings are then computed to generate the graph structure $S$. To enhance discriminability and enforce biologically plausible structures, we apply sparsity constraints and align $S$ with the corresponding PCC-based correlation matrix $A$: 
\begin{equation}
\mathcal{L}_{\mathrm{str}}
=
\min_{S_i}
\;
\sum_{i=1}^{N}
\Big(
\lVert S_i \rVert_{1}
+
\lambda_{\mathrm{VC}} \, \lVert S_i - A_i \rVert_{F}^{2}
\Big),
\label{eq:graph_loss}
\end{equation}
where $N$ denotes the number of patients. This yields a sparse graph structure as a structure view that captures individual brain network topology. Additionally, each subject has a structured clinical text. We encode these texts into high-level semantic embeddings as text view via a pretrained large language model~\cite{Xu2026}.
Based on both structural and text views, we capture subject similarities from complementary perspectives and obtain a fused similarity matrix via similarity network fusion (SNF)~\cite{Wang2014}.
\subsection{Subtype Prototype Discovery}

With the fused similarity matrix, a $k$-nearest neighbor graph is then constructed, and unsupervised spectral clustering is applied to partition patients into latent subtypes. To capture shared brain network patterns within each subtype, we construct a subtype prototype graph for each subtype, serving as the guidance for subsequent contrastive learning. Specifically, each sample's brain functional graph is represented as $G_i \in \mathbb{R}^{M \times D}$, with $D$ denoting the node feature dimension, and let $\{ G_i \}_{i=1}^{N_k}$ indicate the sample set for the $k$-th subtype, where $N_k$ denotes the number of samples in the $k$-th subtype. We stack the subtype's sample graphs along the sample dimension to produce  $\mathbf{G}^{(k)} = [ G_1, G_2, \ldots, G_{N_k} ] \in \mathbb{R}^{N_k \times M \times D}$. A dual-level attention mechanism is applied to model subtype brain networks: node-level self-attention $A_{\mathrm{node}}(\cdot)$ is proposed to capture ROI-level connectivity dependencies across brain regions, while sample-level self-attention $A_{\mathrm{sub}}(\cdot)$ is designed to reflect each sample's contribution to the subtype's shared structure. Attention outputs are $\mathbf{G}_{\mathrm{node}}^{(k)} = A_{\mathrm{node}}\!\left(\mathbf{G}^{(k)}\right)$, $\mathbf{G}_{\mathrm{sub}}^{(k)} = A_{\mathrm{sub}}\!\left(\mathbf{G}_{\mathrm{node}}^{(k)}\right)$, where $\mathbf{G}_{\mathrm{node}}^{(k)} \in \mathbb{R}^{N_k \times M \times D},\ \mathbf{G}_{\mathrm{sub}}^{(k)} \in \mathbb{R}^{N_k \times M \times D}$. Finally, the attention-weighted sample graphs are averaged along the sample dimension to obtain the prototype graph for the $k$-th subtype: $H^{(k)} = \frac{1}{N_k} \sum_{i=1}^{N_k} \mathbf{G}_{\mathrm{sub}}^{(k,i)}$, where $\mathbf{G}_{\mathrm{sub}}^{(k,i)} \in \mathbb{R}^{M \times D}$ represents the $i$-th patient's graph of $k$-th subtype after sample-level attention. The prototype graph captures stable subtype-level common structures across all samples within the subtype, providing a reliable intra-subtype structural reference for subsequent subtype-guided contrastive learning.

\subsection{Subtype-Guided Contrastive Learning}
The subtype-guided contrastive learning strategy leverages subtype prototype graphs as structural priors to enhance intra-subtype consistency and between-class separability.
Dynamic queues of controls and patients for covering a rich set of samples from preceding batches are constructed to ensure embedding consistency across the samples.
In our study, BrainNetCNN~\cite{Kawahara2017} is chosen as the encoder and momentum encoder~\cite{He2020} for extracting the graph embedding, $g_i$ and $g_i^m$.
For each sample embedding $g_i$, the positive pair is defined as the prototype embedding $h^{\mathbb{I}(g_i)}$ of its corresponding subtype prototype graph, while the negative pairs are drawn from opposite-label queue embeddings $Q_{\neg y_i}$ and the opposite-label subtype prototype graph embeddings $H_{\neg y_i}$, where $\neg y_i$ denotes the opposite label of sample $i$.
The subtype-guided contrastive loss $L_{con}$ encourages each sample embedding $g_i$ to be close to its positive sample, while being pushed away from negative samples, enforcing intra-subtype consistency.
Moreover, a hard negative mining scheme is used for focusing on confusing cross-class pairs. 
To enforce stable and consistent representation learning, we introduce a consistency regularization $\mathcal{L}_{\mathrm{cr}}$ that penalizes the difference between the embedding $g_i$ from the encoder and the corresponding embedding $g_i^m$ from the momentum encoder.
Combined with cross-entropy loss $\mathcal{L}_{\mathrm{cls}}$, the overall objective function is formulated as:
\begin{equation}
\label{eq:total_loss_bce}
\begin{aligned}
\mathcal{L}_{\mathrm{total}}
=\min \frac{1}{N+T} \sum_{i=1}^{N+T} \Bigg\{ \;
 - \Big(
y_i \log \hat{y}_i + (1-y_i) \log(1-\hat{y}_i)
\Big)
 + \lambda_{\mathrm{cr}} \, \| g_i - g_i^m \|_2^2
 \\
+ \lambda_{\mathrm{con}}
\Big[
-\log
\frac{\exp(\mathrm{sim}(g_i, h^{\mathbb{I}(g_i)}) / \tau)}
{\exp(\mathrm{sim}(g_i, h^{\mathbb{I}(g_i)}) / \tau)
+ \sum_{x \in Q_{\neg y_i} \cup H_{\neg y_i}}
\exp(\mathrm{sim}(g_i, x)/\tau)}
\Big]
\Bigg\},
\end{aligned}
\end{equation}
where $T$ denotes the number of healthy controls, $\tau$ is the temperature parameter controlling the sharpness of the contrastive distribution, $\lambda_{\mathrm{con}}$ and $\lambda_{\mathrm{cr}}$ balance the contributions of subtype-guided contrastive learning and consistency regularization.

\section{Experiments and Results}
\subsection{Datasets and Experimental Details}
We evaluated our model on the *** dataset\footnote{Approved by the Medical Science Ethics Committee of **** University (Ref. ****).} and the ABIDE dataset, covering three diagnostic tasks (MDD, BD, ASD). The *** dataset was preprocessed with DPABI~\cite{Yan2016} and parcellated into 116 AAL regions. Preprocessing included normalization to MNI space (3 mm), 6 mm Gaussian spatial and temporal smoothing, and 0.01-0.08 Hz band-pass filtering to remove low-frequency drifts and high-frequency noise. It includes 246 HCs (150F/96M, 26.9 $\pm$ 6.19 yrs, 14-51 yrs), 181 MDD (130F/51M, 17 $\pm$ 5 yrs, 12-51 yrs), and 142 BD (101F/41M, 17 $\pm$ 3.9 yrs, 12-39 yrs), all scanned at a single site with consistent protocols. ABIDE, a public ASD dataset, includes 340 ASD (45F/295M, 16.72 $\pm$ 7.24 yrs, 7-50 yrs) and 468 HCs (90F/378M, 16.84 $\pm$ 7.24 yrs, 6-56 yrs) from 20 sites. We used AAL-based pre-processed functional data for evaluation.

\subsection{Classification Results}
To evaluate overall diagnostic performance, we used five-fold cross-validation and compared our method against state-of-the-art approaches using ACC, AUC, SEN, and SPEC. The compared methods include RGTNet~\cite{Wang2024}, GroupINN~\cite{Yan2019}, ASD-DiagNet~\cite{Eslami2019}, MVS-GCN~\cite{Wen2022}, ST-GCN~\cite{Gadgil2020}, BrainTGL~\cite{Liu2023}, BrainGSL~\cite{Wen2023}, DSAM~\cite{Thapaliya2025} and SDBD~\cite{Jin2024}.

\begin{table}[htbp]
\caption{Classification performance comparison with state-of-the-art methods. 
BrainSCL-s indicate the supervised learning baseline with an encoder, trained without contrastive learning or subtype aggregation;
BrainSCL-cl represents traditional supervised contrastive learning;
BrainSCL-m constructs subtype prototype graphs by averaging samples within each subtype;
$K$ in BrainSCL indicates the number of subtypes;
BrainSCL-t and BrainSCL-g denote single-view subtype clustering variants using only clinical text semantics and only brain graph structures, respectively.
}\label{tab:results}
\small 
\resizebox{\textwidth}{!}
{ 
\begin{tabular}{l|*{4}{c}|*{4}{c}|*{4}{c}} 
\hline
Methods & \multicolumn{4}{c|}{HC vs. BD} & \multicolumn{4}{c|}{HC vs. MDD} & \multicolumn{4}{c}{HC vs. ASD} \\
 & ACC & AUC & SEN & SPEC & ACC & AUC & SEN & SPEC & ACC & AUC & SEN & SPEC \\
\hline
RGTNet & 67.1 & 71.2 & 53.7 & 75.9 & 63.9 & 65.6 & 44.0 & 77.9 & 62.0 & 66.1 & 63.7 & 60.1 \\
GroupINN & 67.9 & 63.3 & 62.8 & 67.1 & 66.8 & 65.3 & 63.1 & 65.8 & 63.9 & 63.2 & 61.5 & 57.4 \\
ASD-DiagNet & 73.3 & 70.1 & 58.5 & 79.6 & 68.2 & 66.7 & 60.2 & 74.3 & 68.3 & 67.8 & 60.3 & 67.8 \\
MVS-GCN & 66.9 & 64.2 & 62.1 & 73.5 & 68.3 & 68.2 & 63.8 & 61.2 & 69.9 & 69.1 & \textbf{70.2} & 63.1 \\
ST-GCN & 67.1 & 57.5 & 56.6 & 73.5 & 58.1 & 52.3 & 53.3 & 55.1 & 57.3 & 51.7 & 54.8 & 48.9 \\
BrainTGL & 72.0 & 70.8 & \textbf{76.1} & 73.9 & 73.2 & 68.9 & 70.2 & 78.0 & 67.8 & 67.6 & 62.2 & 72.9 \\
BrainGSL & 73.6 & 69.5 & 59.2 & 79.9 & 72.9 & 72.3 & 59.5 & 81.9 & 67.3 & 66.9 & 64.4 & 69.4 \\
DSAM & 75.4 & 75.3 & 63.7 & 82.4 & 74.2 & 73.3 & \textbf{71.5} & 78.6 & 67.6 & 65.4 & 62.4 & 66.9 \\
SDBD & 76.0 & 71.8 & 60.8 & 79.2 & 74.8 & 73.3 & 60.4 & 72.2 & 70.4 & 69.3 & 62.0 & 71.9 \\
\hline 
BrainSCL-s & 73.3 & 71.7 & 64.1 & 79.3 & 69.4 & 68.4 & 61.5 & 75.2 & 65.6 & 64.2 & 52.5 & 69.3 \\
BrainSCL-cl & 75.7 & 79.3 & 63.3 & 82.9 & 72.6 & 79.0 & 62.8 & 79.8 & 70.0 & 74.7 & 58.6 & 78.0 \\
BrainSCL-m & 76.0 & 78.6 & 66.0 & 81.3 & 75.8 & 78.0 & 67.2 & 80.2 & 69.7 & 74.9 & 60.9 & 76.3 \\
BrainSCL-t & 75.5 & 81.0 & 60.6 & 84.0 & 72.6 & 79.1 & 66.5 & 77.7 & 70.3 & 74.9 & 59.8 & 77.9 \\
BrainSCL-g & 75.5 & 80.2 & 61.3 & 83.7 & 75.6 & 80.6 & 64.1 & 84.1 & 70.1 & 73.9 & 62.5 & 75.8 \\
\textbf{BrainSCL(K=3)} & \textbf{77.8} & \textbf{81.5} & {66.9} & 84.1 & \textbf{76.8} & \textbf{82.1} & {68.1} & 83.3 & \textbf{71.3} & \textbf{75.9} & {61.9} & \textbf{78.2} \\
\hline 
BrainSCL(K=2) & 76.5 & 79.9 & 59.2 & \textbf{86.6} & 76.1 & 80.0 & 65.6 & \textbf{84.5} & 69.9 & 74.2 & 60.7 & 76.7 \\
BrainSCL(K=4) & 76.3 & 80.1 & 59.7 & 85.7 & 74.5 & 79.0 & 67.4 & 79.9 & 69.5 & 73.5 & 58.6 & 77.5 \\
\hline
\end{tabular}}
\end{table}

\subsubsection{Comparison with SOTA} Our model achieved the best performance on MDD, BD, and ASD diagnosis (Table~\ref{tab:results}), with ACC of 76.8\% for MDD, 77.8\% for BD, and 71.3\% for ASD, along with improved AUCs compared to competing methods.
These results demonstrate the advantage of our method against other competing methods on various disorder diagnosis tasks.

\subsubsection{Ablation Study} We evaluated the contribution of key modules on the *** dataset (MDD and BD) and ABIDE using stepwise ablation with identical data splits and training settings. BrainSCL-s showed the lowest performance, while BrainSCL-cl achieved only limited improvement.
BrainSCL-m is inferior to our full method, demonstrating that precise subtype prototypes are crucial for contrastive learning.
Our method yields the best results, indicating that accurately modeling intra-subtype brain network structures provides more reliable guidance for contrastive learning.
In addition, multi-view subtype discovery consistently outperforms single-view clustering, demonstrating that multiple views provide a more comprehensive representation for subtype clustering and prototype construction, ultimately facilitate the subtype-guided contrastive learning.
We also analyze the effect of subtype granularity by varying the subtype number and  observe that $K=3$ achieves the best performance, while coarser or finer subtype partitions led to inferior results. This indicates that an appropriate cluster partition is critical for accurate prototype construction and contrastive learning.

\subsection{Interpretability Analysis}
\subsubsection{Validity of Latent Subtype Discovery} 
\begin{figure}[htbp]
    \centering

    \parbox{0.32\textwidth}{
        \centering
        \includegraphics[width=\linewidth]{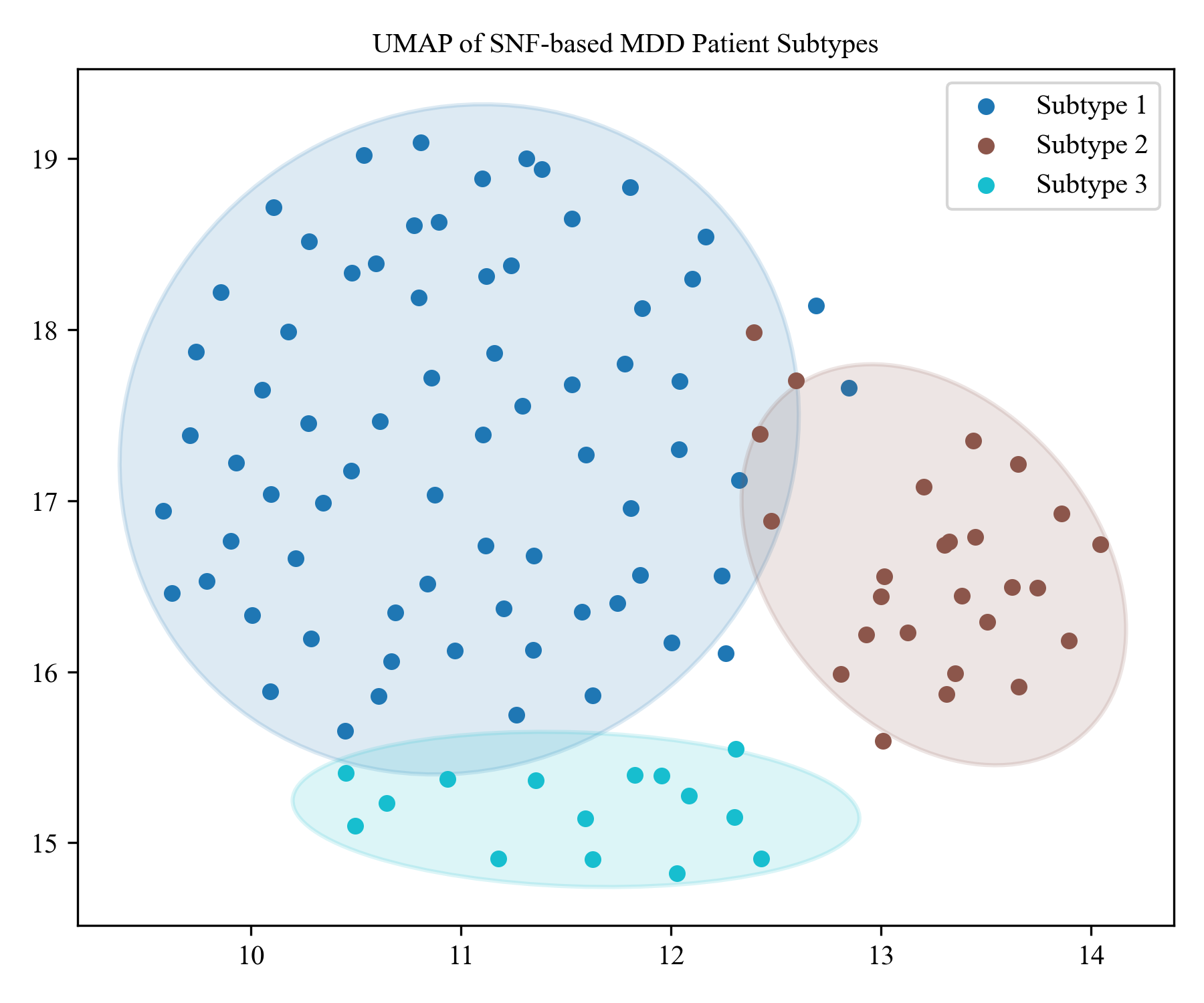}
    }
    \hfill
    \parbox{0.32\textwidth}{
        \centering
        \includegraphics[width=\linewidth]{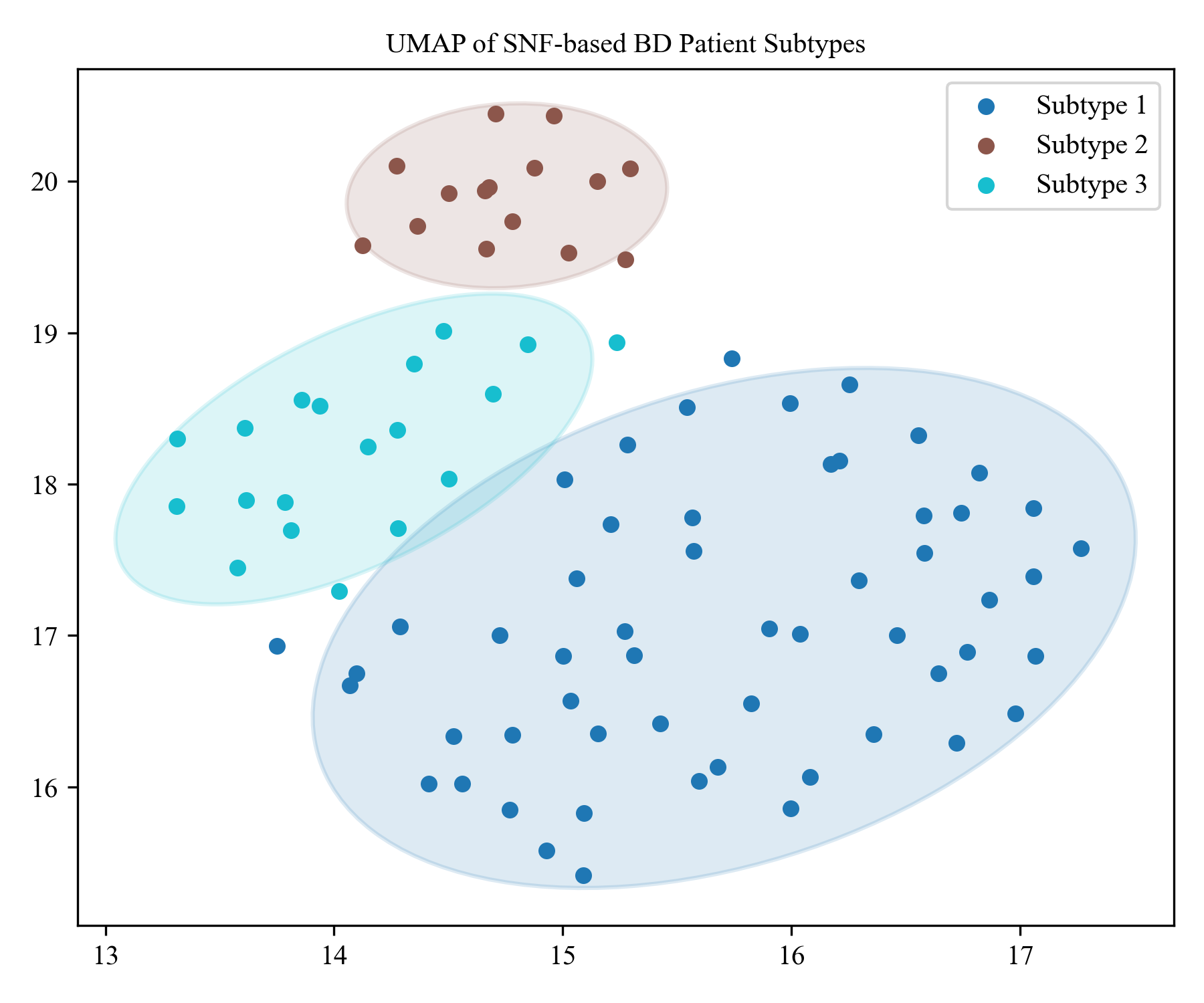}
    }
    \hfill
    \parbox{0.32\textwidth}{
        \centering
        \includegraphics[width=\linewidth]{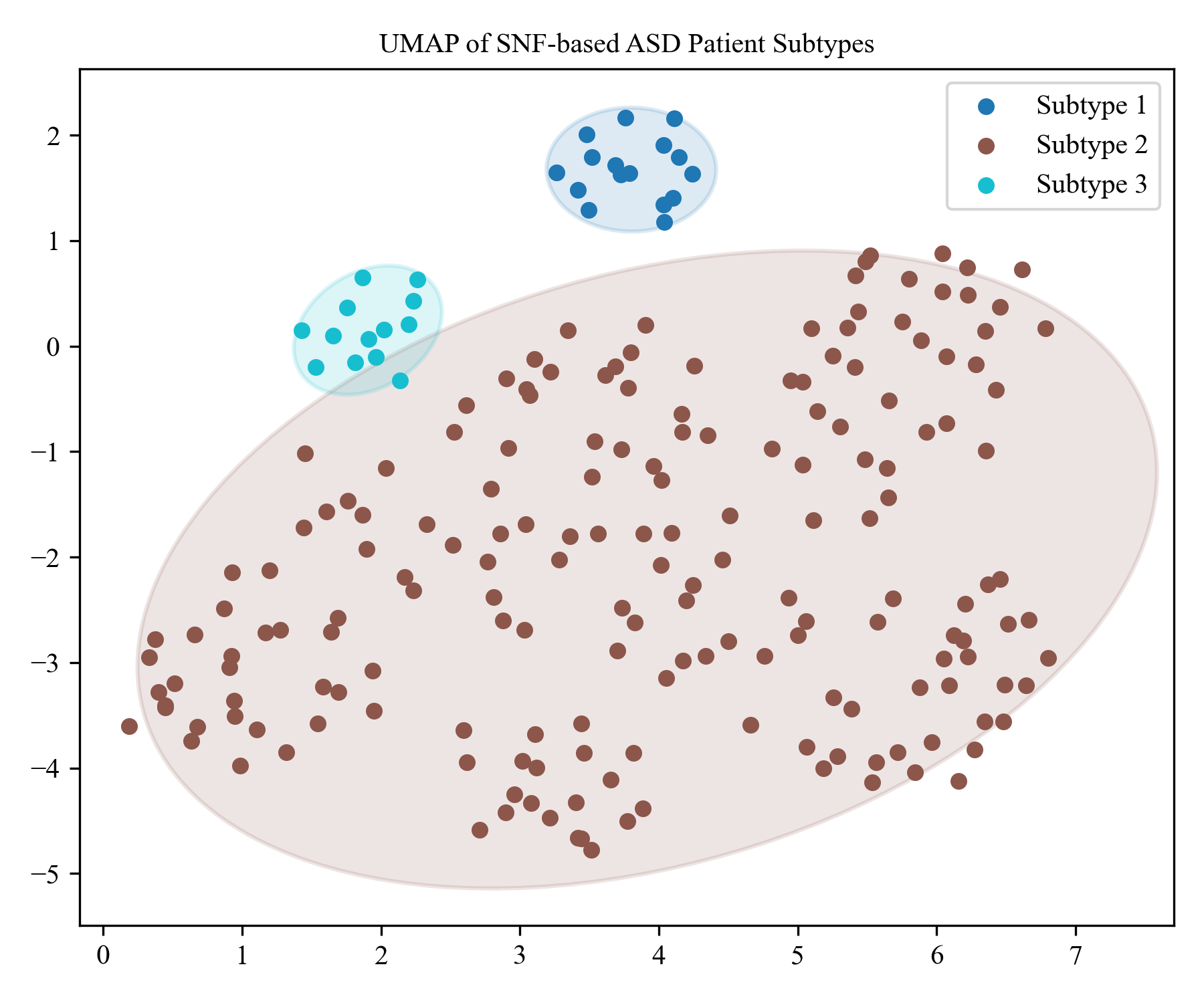}
    }

    \medskip

    \caption{SNF-based patient subtypes visualized on the three disorder diagnosis scenarios with UMAP~\cite{McInnes2018}.
    }
    \label{fig4}
\end{figure}

To assess the validity  of the identified patient subtypes, we qualitatively evaluated the clustering on the three disorder diagnosis scenarios (MDD, BD and ASD) using low-dimensional visualization 
(Fig.~\ref{fig4}). 
Subtypes were largely well-separated with compact intra-subtype distributions, indicating that the fused similarity matrix preserves subtype-related structure in the low-dimensional space. At subtype boundaries, a few samples overlap, suggesting a continuous rather than strictly separable distribution. 

\subsubsection{Subtype Prototype Graph Interpretability Analysis} 
\begin{figure}[htbp]
    \centering
    \parbox{0.32\textwidth}{
        \centering
        \includegraphics[width=\linewidth]{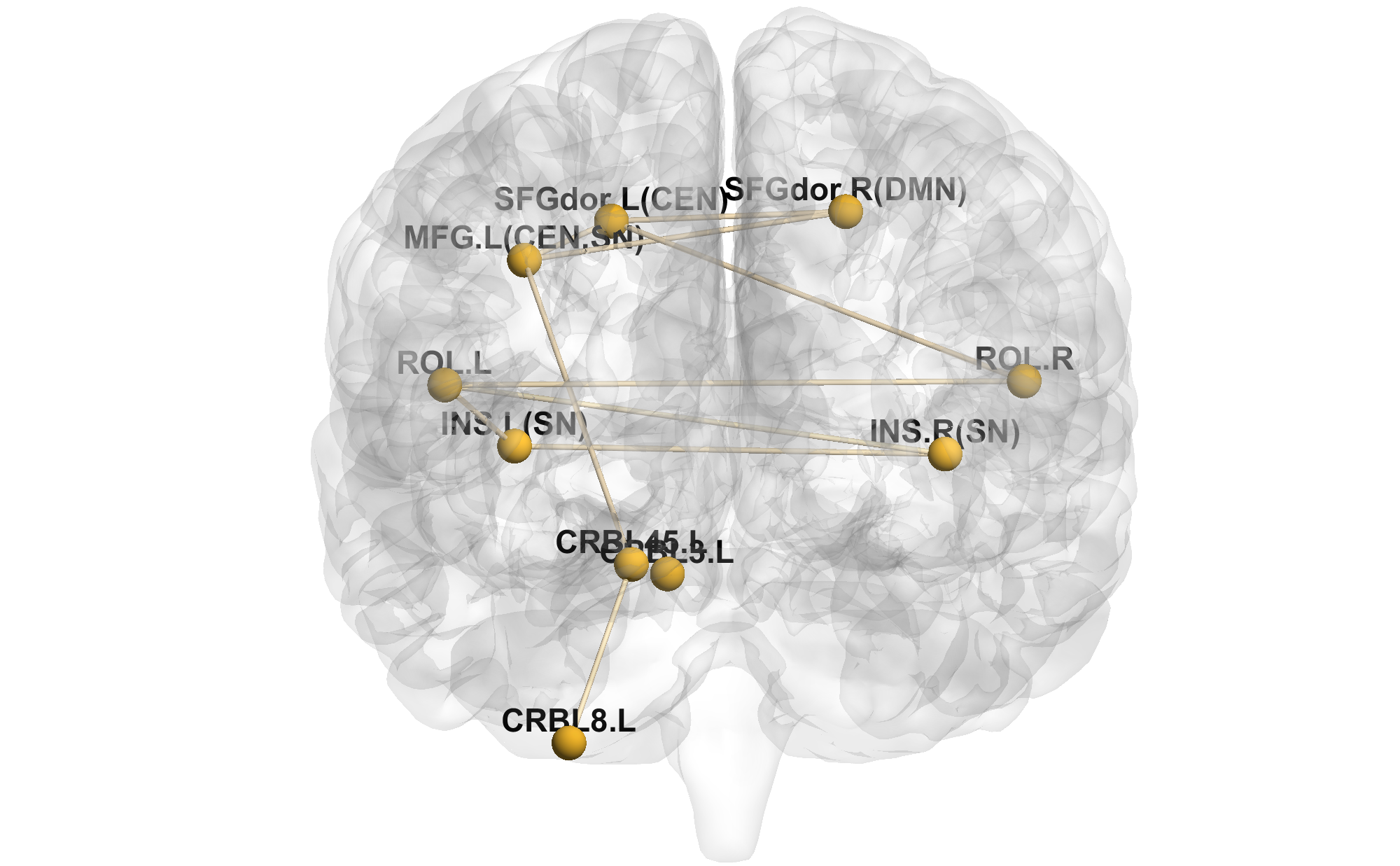}
        {\footnotesize (a) Subtype 1 - MDD}
    }
    \hfill
    \parbox{0.32\textwidth}{
        \centering
        \includegraphics[width=\linewidth]{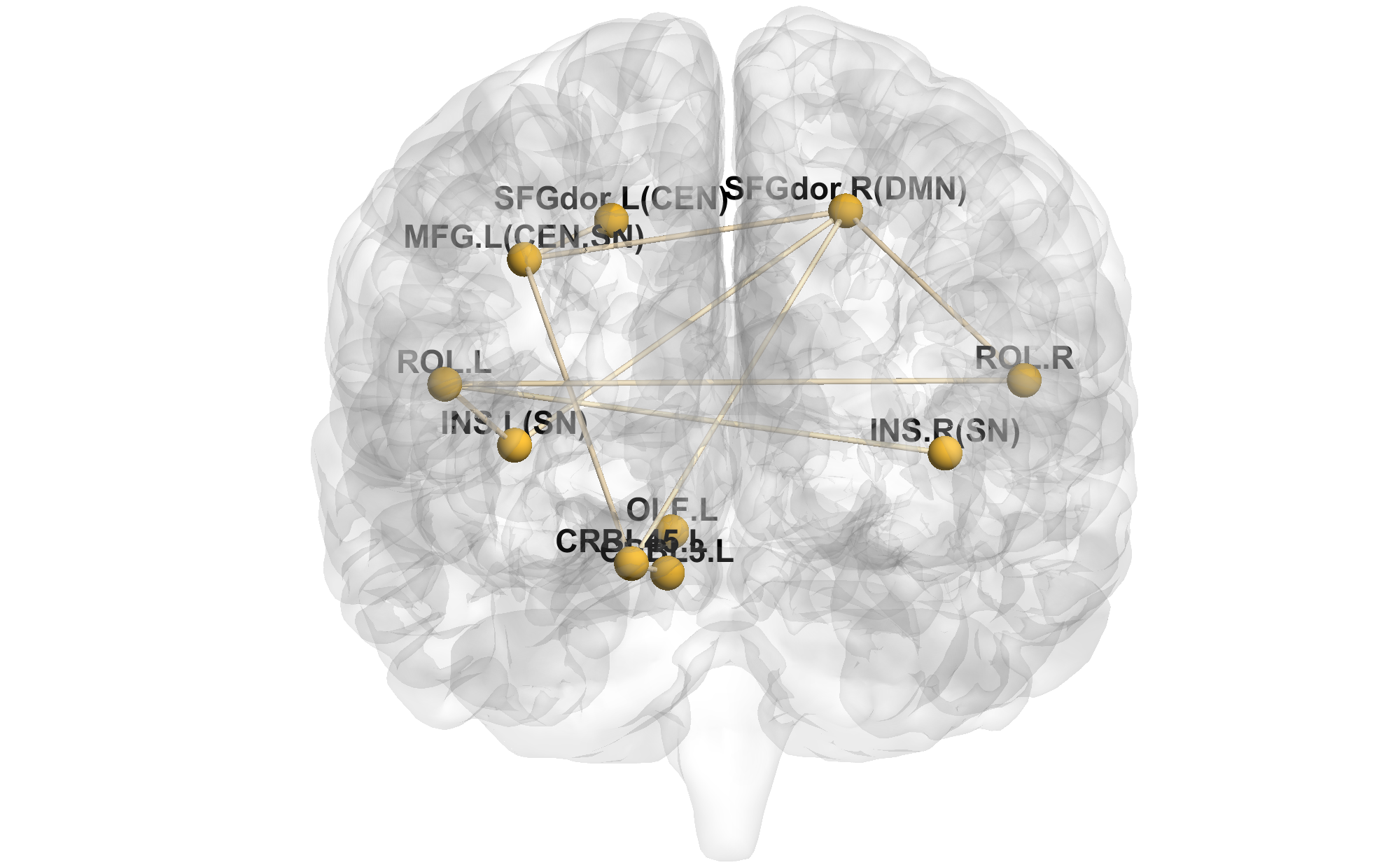}
        {\footnotesize (b) Subtype 2 - MDD}
        
    }
    \hfill
    \parbox{0.32\textwidth}{
        \centering
        \includegraphics[width=\linewidth]{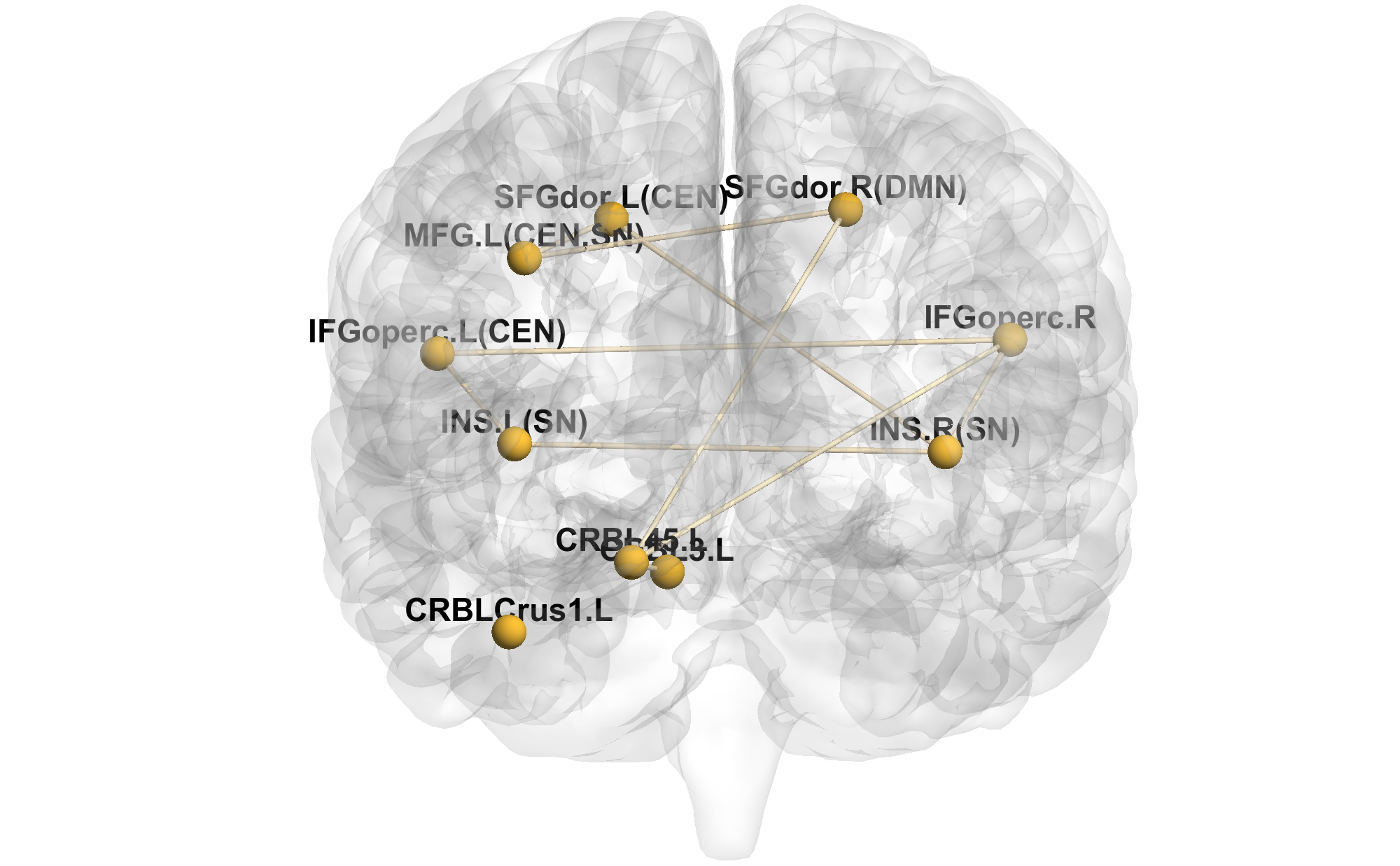}
        {\footnotesize (c) Subtype 3 - MDD}
    }

    \medskip

    \parbox{0.32\textwidth}{
        \centering
        \includegraphics[width=\linewidth]{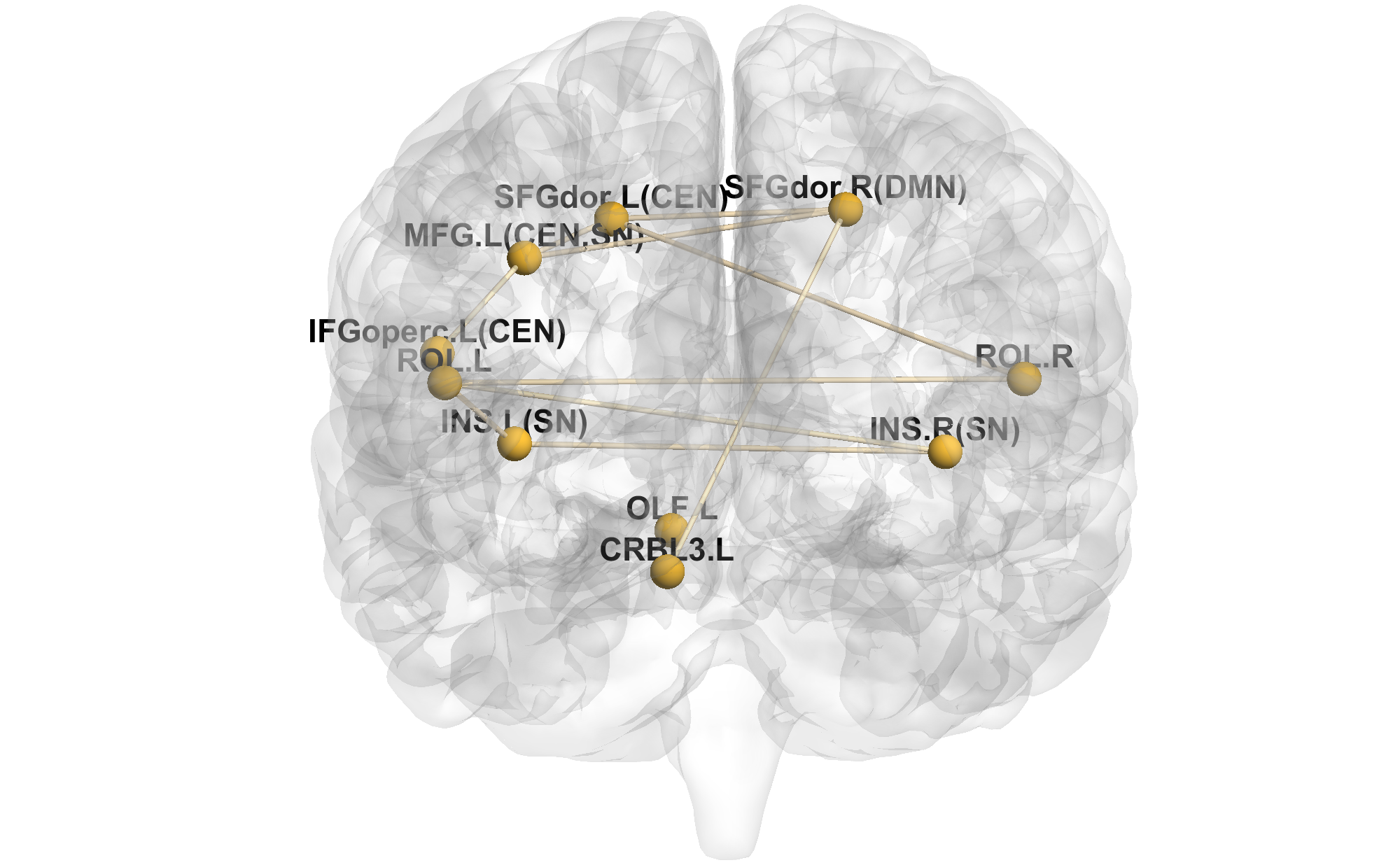}
        {\footnotesize (d) Subtype 1 - BD}
    }
    \hfill
    \parbox{0.32\textwidth}{
        \centering
        \includegraphics[width=\linewidth]{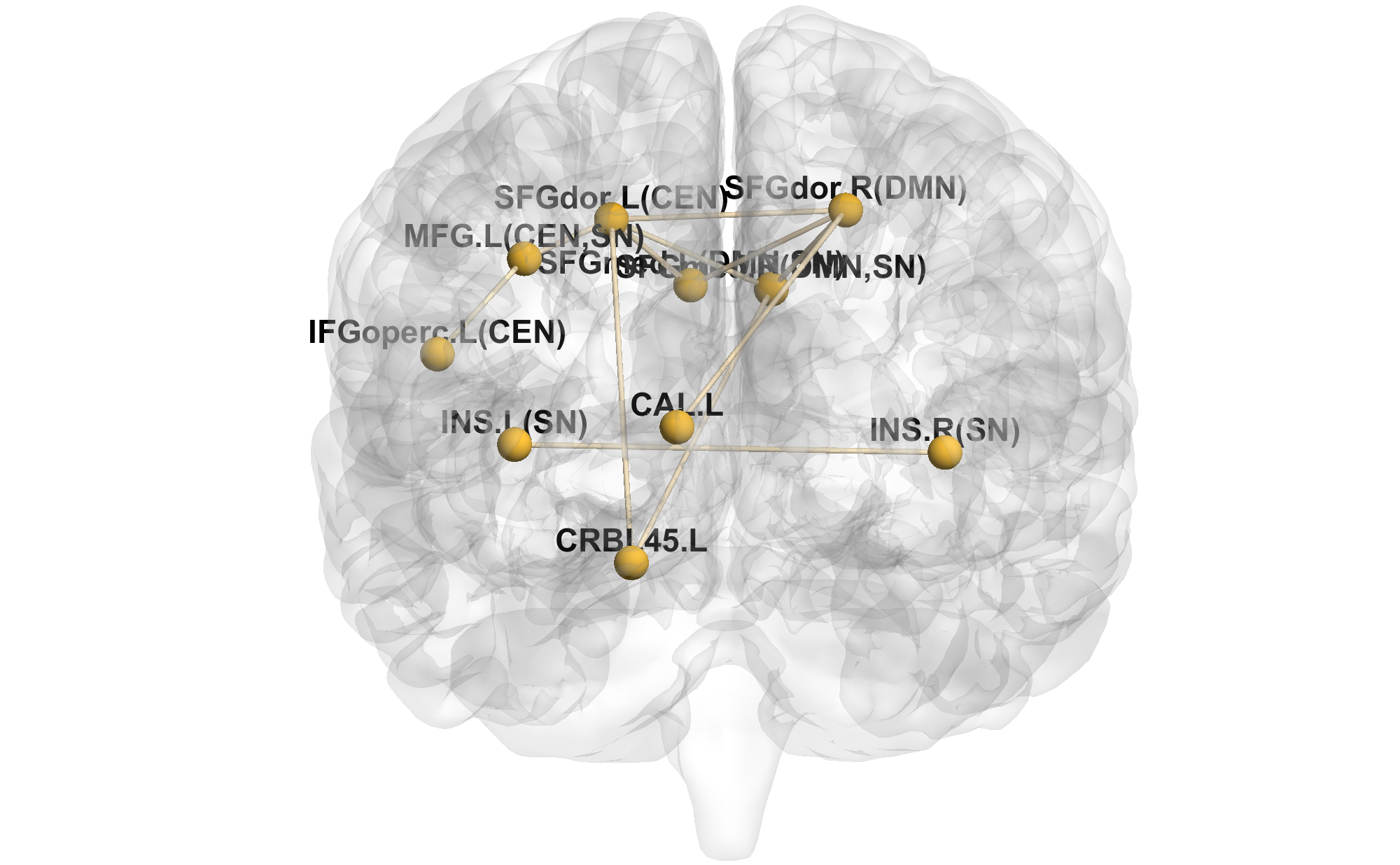}
        {\footnotesize (e) Subtype 2 - BD}
    }
    \hfill
    \parbox{0.32\textwidth}{
        \centering
        \includegraphics[width=\linewidth]{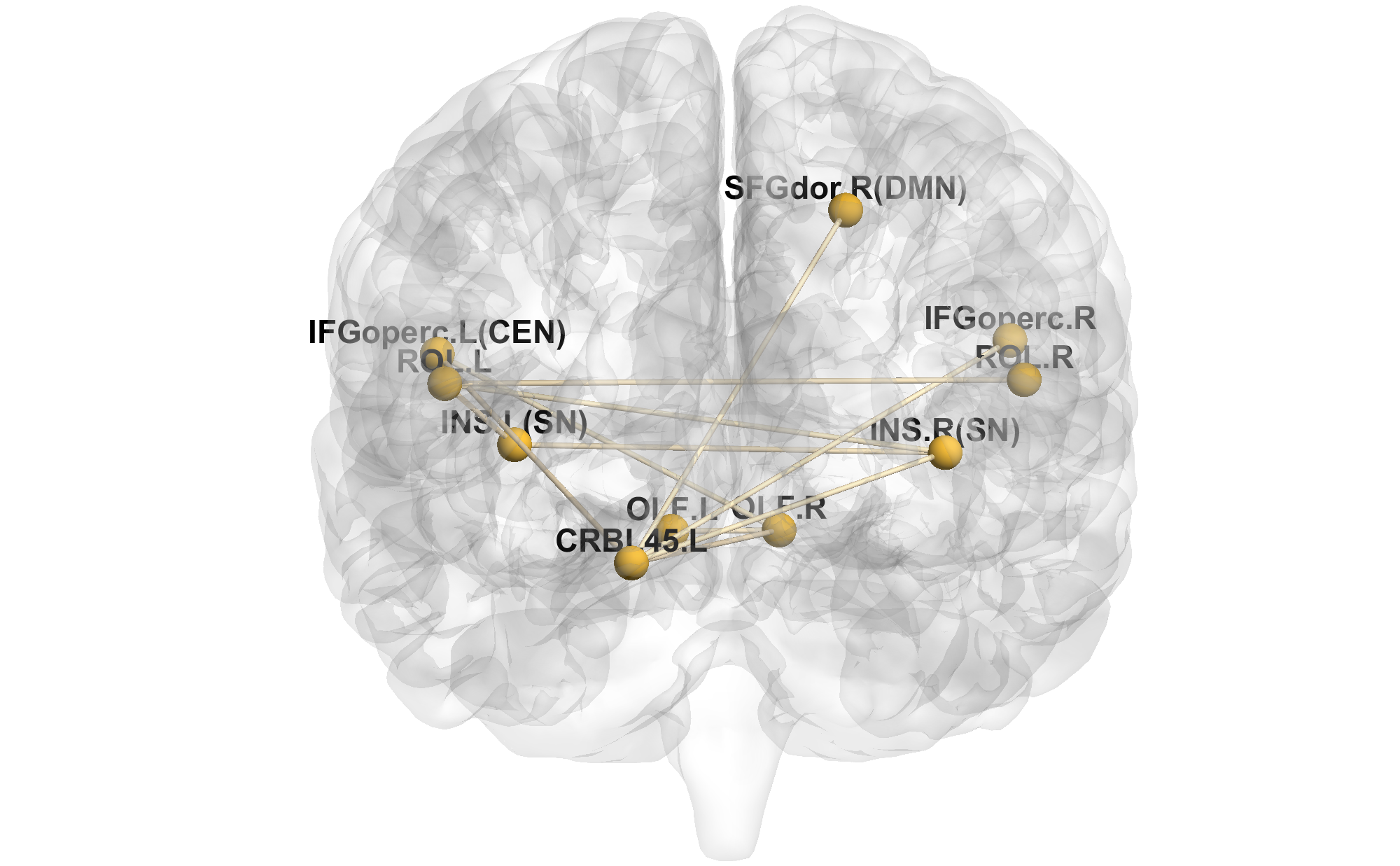}
        {\footnotesize (f) Subtype 3 - BD}
    }

    \caption{Top 10 brain regions in subtype prototype graphs for three subtypes. 
    }
    \label{fig5}
\end{figure}

Taking MDD and BD as representative disorders, we analyze the interpretability of the learned subtype prototype graphs by analyzing the functional network distribution of the top-ranked regions in the subtype prototype graphs (Fig.~\ref{fig5}). Across all three identified subtypes, the top-ranked regions are consistently concentrated in the Salience Network (SN) and Central Executive Network (CEN), with additional involvement of the Default Mode Network (DMN).
Notably, several core regions, including SFGdor.R (DMN), INS.R (SN), and INS.L (SN), consistently emerge across all subtype prototype graphs, indicating the existence of shared cross-subtype neural hubs that may reflect common pathological substrates.
These results suggest that the proposed subtype discovery and aggregation mechanism captures neurobiologically meaningful structures.

\section{Conclusion}

Psychiatric populations exhibit high heterogeneity,
leading to unstable brain network representations and limited diagnostic performance. This heterogeneity poses a significant challenge for contrastive learning, particularly in constructing semantically reliable positive pairs.
To address this, we propose a subtype-guided contrastive learning framework (BrainSCL) that discovers latent patient subtypes, constructs subtype prototype graphs, and leverages subtype-guided contrastive learning for discriminative representation learning.
This framework improves intra-subtype consistency and between-class separability in the learned representations. Experiments show our method outperforms state-of-the-art approaches across multiple psychiatric diagnosis tasks, while ablation studies confirm the effectiveness of subtype guidance, subtype aggregation, and the contrastive learning strategy.
Interpretability analyses validate the discovered subtypes and show that subtype prototype graphs capture neurobiologically meaningful, disease-relevant network structures.

\end{document}